\documentclass[runningheads]{llncs}

\usepackage{eccv}

\usepackage{eccvabbrv}

\usepackage{graphicx}
\usepackage{booktabs}
\usepackage[ruled,linesnumbered]{algorithm2e}
\usepackage{utfsym}
\usepackage{multirow}
\usepackage{booktabs}
\usepackage{makecell}
\usepackage{ulem}
\usepackage{adjustbox}
\usepackage{wrapfig}
\usepackage{bbding}
\renewcommand\footnotemark{}
\useunder{\uline}{\ul}{}

\usepackage[accsupp]{axessibility}  

\usepackage{hyperref}

\usepackage{orcidlink}

\begin{document}

\title{Control-A-Video: Controllable Text-to-Video Diffusion Models with Motion Prior and Reward Feedback Learning}

\author{Weifeng Chen\thanks{* ~ Contributed equally}\textsuperscript{*}, \qquad
Yatai Ji\textsuperscript{*}, \qquad
Jie Wu, \qquad 
Hefeng Wu\thanks{\textsuperscript{\Envelope} ~ Corresponding author}\Envelope, \qquad
Pan Xie, \\ Xin Xia,\qquad Jiashi Li,\qquad Xuefeng Xiao,\qquad Liang Lin\\
\tt Project Page: \textcolor{blue}{https://controlavideo.github.io}}
\institute{}

\maketitle



\begin{abstract}

Recent advances in text-to-image (T2I) diffusion models have enabled impressive image generation capabilities guided by text prompts. However, extending these techniques to video generation remains challenging, with existing text-to-video (T2V) methods often struggling to produce high-quality and motion-consistent videos. In this work, we introduce Control-A-Video, a controllable T2V diffusion model that can generate videos conditioned on text prompts and reference control maps like edge and depth maps. To tackle video quality and motion consistency issues, we propose novel strategies to incorporate content prior and motion prior into the diffusion-based generation process. Specifically, we employ a first-frame condition scheme to transfer video generation from the image domain. Additionally, we introduce residual-based and optical flow-based noise initialization to infuse motion priors from reference videos, promoting relevance among frame latents for reduced flickering. Furthermore, we present a Spatio-Temporal Reward Feedback Learning (ST-ReFL) algorithm that optimizes the video diffusion model using multiple reward models for video quality and motion consistency, leading to superior outputs. Comprehensive experiments demonstrate that our framework generates higher-quality, more consistent videos compared to existing state-of-the-art methods in controllable text-to-video generation. 

\end{abstract}

\section{Introduction}
\label{sec:intro}

In recent years, the field of text-based visual content generation has witnessed rapid growth.  Current T2I diffusion models \cite{Ramesh_Dhariwal_Nichol_Chu_Chen,saharia2022photorealistic, Rombach_Blattmann_Lorenz_Esser_Ommer_2022} trained on large-scale image-text pairs, showcase impressive capabilities in producing high-quality images guided by user-provided text prompts. Built on these pre-trained T2I models, personalized generation \cite{Ruiz_Li_Jampani_Pritch_Rubinstein_Aberman_2022, gal2022image} and conditional generation \cite{zhang2023adding,mou2023t2i} provide more fine-grained control over the generated image. Further advancements in T2I models have paved the way for personalized generation\cite{Ruiz_Li_Jampani_Pritch_Rubinstein_Aberman_2022, gal2022image} and conditional generation\cite{zhang2023adding, mou2023t2i}, offering finer control over generated videos. The success achieved in image generation has been seamlessly extended to video generation, where Text-to-Video (T2V) diffusion models\cite{singer2022make, ho2022imagen} excel in generating coherent videos driven by text prompts. While some approaches leverage T2I models for one-shot\cite{wu2022tune} or zero-shot\cite{khachatryan2023text2video, qi2023fatezero} video generation, the outcomes often lack consistency or diversity. Other methods\cite{esser2023structure, videocomposer, followyourpose} introduce conditional video generation with control maps, mirroring image-based techniques. 

However, these approaches still grapple with the challenge of producing high-quality and motion-consistent videos.
Concerning video quality, we analyze it from two distinct perspectives: technical quality, characterized by fewer artifacts and reduced blur, and aesthetic quality, gauged subjectively based on human perception of visual appeal. Motion consistency involves both object consistency (where subjects and backgrounds remain unchanged between frames) and motion smoothness (where motion adheres to the principles of physics and no flickering). 

\begin{figure}[t]
\centering
  \includegraphics[width=0.95\linewidth]{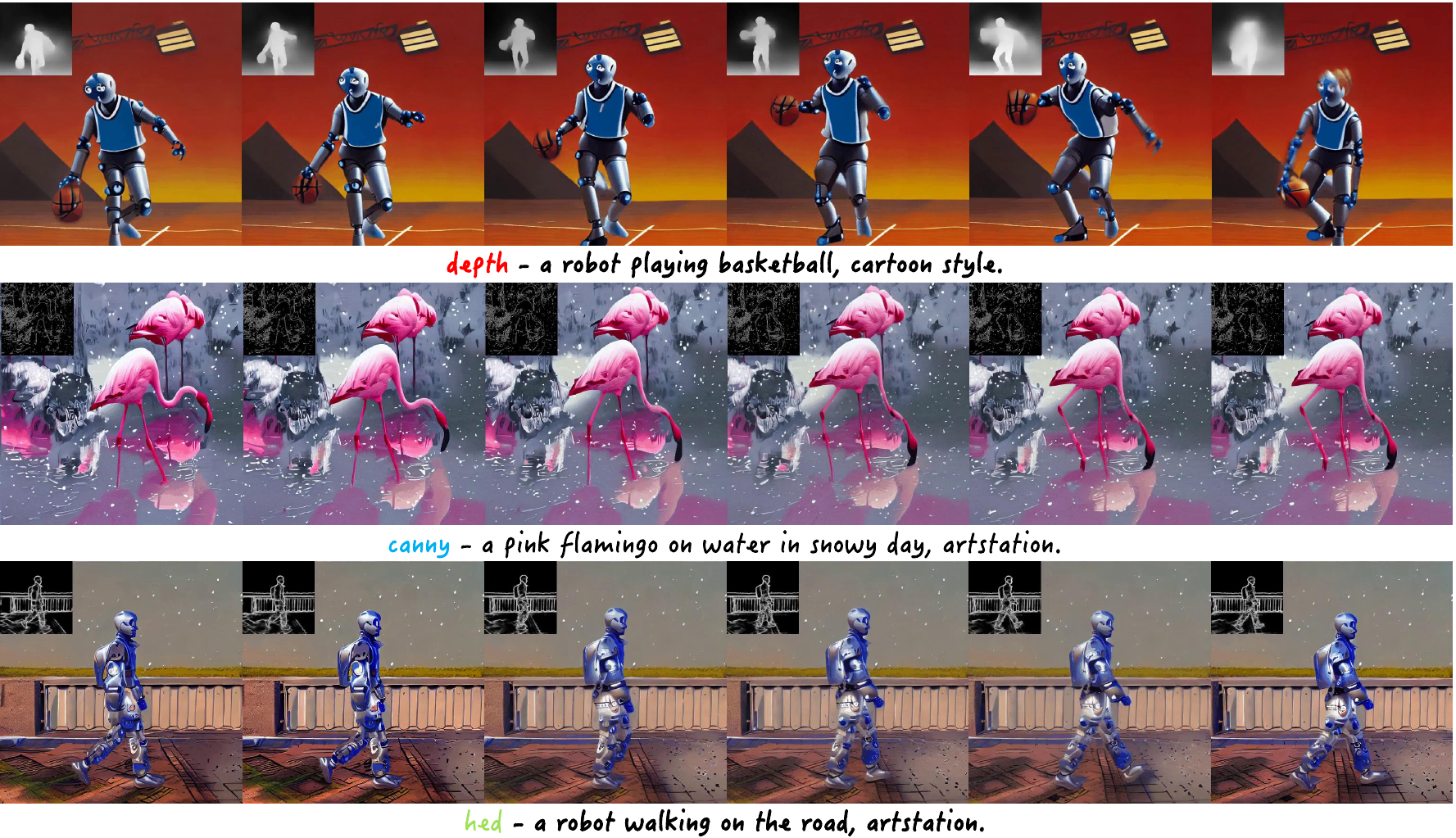}
\caption{Our model generates high-quality and consistent videos conditioned on a text prompt and additional control maps, such as \textcolor{red}{depth maps} (first row), \textcolor{cyan}{canny edge maps} (second row), \textcolor{green}{hed edge maps} (third row). }
\label{fig:head} 
\vspace{-0.5cm}
\end{figure}%

In response to the aforementioned challenges, we presents a controllable T2V model, namely Control-A-Video, capable of generating based on text and reference control maps, such as edge, depth maps, as shown in Figure \ref{fig:head}. 
We develop our video generative model by reorganizing a pretrained controllable T2I model \cite{zhang2023adding}, incorporating additional trainable temporal layers, and utilizing a spatial-temporal self-attention mechanism that facilitates fine-grained interactions between frames. This meticulous architectural design, coupled with the incorporation of control signals, notably preserves motion smoothness. 
Furthermore, we propose novel strategies to introduce content prior and motion prior for diffusion-based generation, boosting motion consistency. Additionally, we introduce ST-ReFL, an algorithm that utilizes multiple reward models to enhance both motion consistency and video quality.

For the content prior, we introduce an training scheme that produces video predicated on the initial frame. With this, it becomes more manageable to disentangle content and temporal modeling. Instead of learning the generation of entire videos, our model focuses on generating subsequent frames with content prior derived from the first frame. This design inherits generative capabilities from the image domain, ensuring consistency with objects in the initial frame.
During inference, we employ a Text-to-Image-to-Image-to-Video (T2I-I2V) pipeline. This involves generating an image and using it as a content prior for video generation.
Meanwhile, another benefit of such strategy is that our model can auto-regressively generate a longer video by treating the last frame of the previous iteration as the initial frame. 

The motion prior come from two types of pioneering noise initialization approaches. 
The first method is to initialize noise based on pixel residual between frames of the source video. 
The second one is to calculate the initial noise of next frame based on optical flow of the source video movement. 
Inspired by \cite{preserve}, these approaches can make the initial distribution of noise latents more reasonable, enhancing relevance among frame latents and improving video consistency. 
By leveraging motion prior, the Control-A-Video is able to closely resemble motion changes in the reference video and produce coherent videos that are less flickering and object consistent. 

While prior contribute to enhancing consistency, the denoising training process encounters challenges in learning to generate high-quality videos. These challenges may manifest as aesthetic issues, artifacts, and object inconsistency, as shown in the lower part of Figure \ref{fig:pipeline}. One potential solution lies in leveraging large-scale and high-quality video datasets. Instead of opting for conventional engineering's approaches, such as manually collecting more data, we present an pioneering solution: introducing ST-ReFL algorithm to enhance the video diffusion model by leveraging diverse reward models can score frame-wise video quality and temporal motion consistency. Utilizing rewards from the reward models, we use ST-ReFL algorithm to optimize the diffusion model. This targeted optimization process is designed to boost the overall reward, leading to enhanced quality and consistency in the model's output. Through the implementation of the ST-ReFL training process, our experiments reveal a noticeable reduction in artifacts, improved aesthetic appeal, and heightened consistency in the generated videos. 

In summary, our contributions can be outlined as follows:
(i) \textbf{Control-A-Video Model}: We introduce Control-A-Video, a controllable Text-to-Video diffusion model designed for video generation based on text prompts and reference control maps.
(ii) \textbf{Content prior}: we propose a novel approach, utilizing the first frame as content prior, making it more manageable to disentangle content and temporal modeling during training. During inference, a T2I-I2V pipeline is employed to transfer text-aligned knowledge from images to videos and enabling auto-regressive generation for extended videos.
(iii) \textbf{Motion prior}: We introduce innovative noise initialization strategies, incorporating both pixel residual-based and optical flow-based approaches. These motion prior, derived from reference videos, significantly enhance the relevance among frame latents, resulting in videos that are more consistent with reduced flickering.
(iv) \textbf{ST-ReFL for Video Diffusion Model}: Addressing inefficiency in denoising training, we propose a ST-ReFL algorithm, which leverages diverse reward models to optimize the trained video diffusion model by scoring its output, leading to improvements in both quality and consistency. To the best of our knowledge, we are the first to introduce feedback learning for optimizing video generation model. (v) Comprehensive experiments showcase that our framework excels in generating higher-quality and consistent videos, achieving state-of-the-art results in controllable text-to-video generation.

\section{Related Work}
\subsection{Text-to-image generation with diffusion models}

Over the past few years, significant advancements have been made in image generation, notably with Denoising Diffusion Probabilistic Models (DDPMs) \cite{Ho_Jain_Abbeel_2020,Song_Meng_Ermon_2020}. DDPMs have surpassed the performance of Generative Adversarial Networks (GANs) \cite{goodfellow2020generative} and Variational Autoencoders (VAEs) \cite{vae2013}. To generate images conditioned on text, several approaches like GLIDE \cite{GLIDE}, DALLE-2 \cite{Ramesh_Dhariwal_Nichol_Chu_Chen}, Imagen \cite{saharia2022photorealistic}, and Latent Diffusion Models (LDMs) \cite{Ramesh_Dhariwal_Nichol_Chu_Chen} have been proposed, training diffusion models on large-scale text-image pairs. For personalized generation, few-shot tuning methods such as Textual Inversion \cite{gal2022image} and Dreambooth \cite{Ruiz_Li_Jampani_Pritch_Rubinstein_Aberman_2022} have been developed. Editing techniques like Prompt2Prompt \cite{hertz2022prompt}, Plug-and-Play \cite{tumanyan2022plug}, and InstructPix2Pix \cite{brooks2022instructpix2pix} offer ways to refine generated images. Notably, ControlNet \cite{zhang2023adding}, T2I-Adapter \cite{mou2023t2i}, and Composer \cite{composer} focus on fine-tuning text-to-image (T2I) models with condition-text-image pairs, enabling the integration of control hints like edges, poses, and depth maps. In this paper, we adopt a similar idea to ControlNet and extend it to video generation with conditions. Recently, Reward Feedback Learning approaches have been introduced in the diffusion domain, such as ImageReward \cite{imagereward}, RAFT \cite{RAFT}, and DPOK \cite{DPOK}, demonstrating a promising path to optimize diffusion models. Our work focuses on utilizing reward models to optimize video diffusion models.

\subsection{Text-to-video generation with diffusion models}

The success of diffusion models in text-to-image (T2I) generation has inspired research into video generation. Notable works like VDM \cite{ho2022video}, Imagen Video \cite{ho2022imagen}, Make-A-Video \cite{singer2022make}, and Animatediff \cite{animatediff} extend T2I diffusion models by training on large text-video datasets. To generate longer videos, LVDM \cite{lvdm}, MCVD \cite{mcvd}, and Align-Your-Latents \cite{alignyourlatent} adopt an auto-regressive approach, sequentially generating frames for temporal coherence.

For text-based video translation, zero-shot methods such as Text2Video-Zero \cite{khachatryan2023text2video}, FateZero \cite{qi2023fatezero}, Vid2VidZero \cite{vid2vid-zero}, and Video-P2P \cite{vid-p2p} explore diffusion model latent spaces and employ temporal attention, but may lack temporal consistency. To improve consistency, flow-based approaches like Render-A-Video \cite{renderavideo} and TokenFlow \cite{tokenflow} introduce flow constraints, while tuning-based methods like Tune-A-Video \cite{wu2022tune} and CodeF \cite{codef} fine-tune models during inference, albeit requiring additional tuning. Gen-1 \cite{esser2023structure} and VideoCompose \cite{videocomposer} train video diffusion models with conditional maps. In contrast, this paper presents a novel two-stage generation approach (T2I-I2V) and employs noise manipulation to enhance temporal consistency in text-based video translation. Besides, we propose to use ST-ReFL to improve video quality and consistency.

\begin{figure*}[!t]
   \centering
   \includegraphics[width=0.98\linewidth]{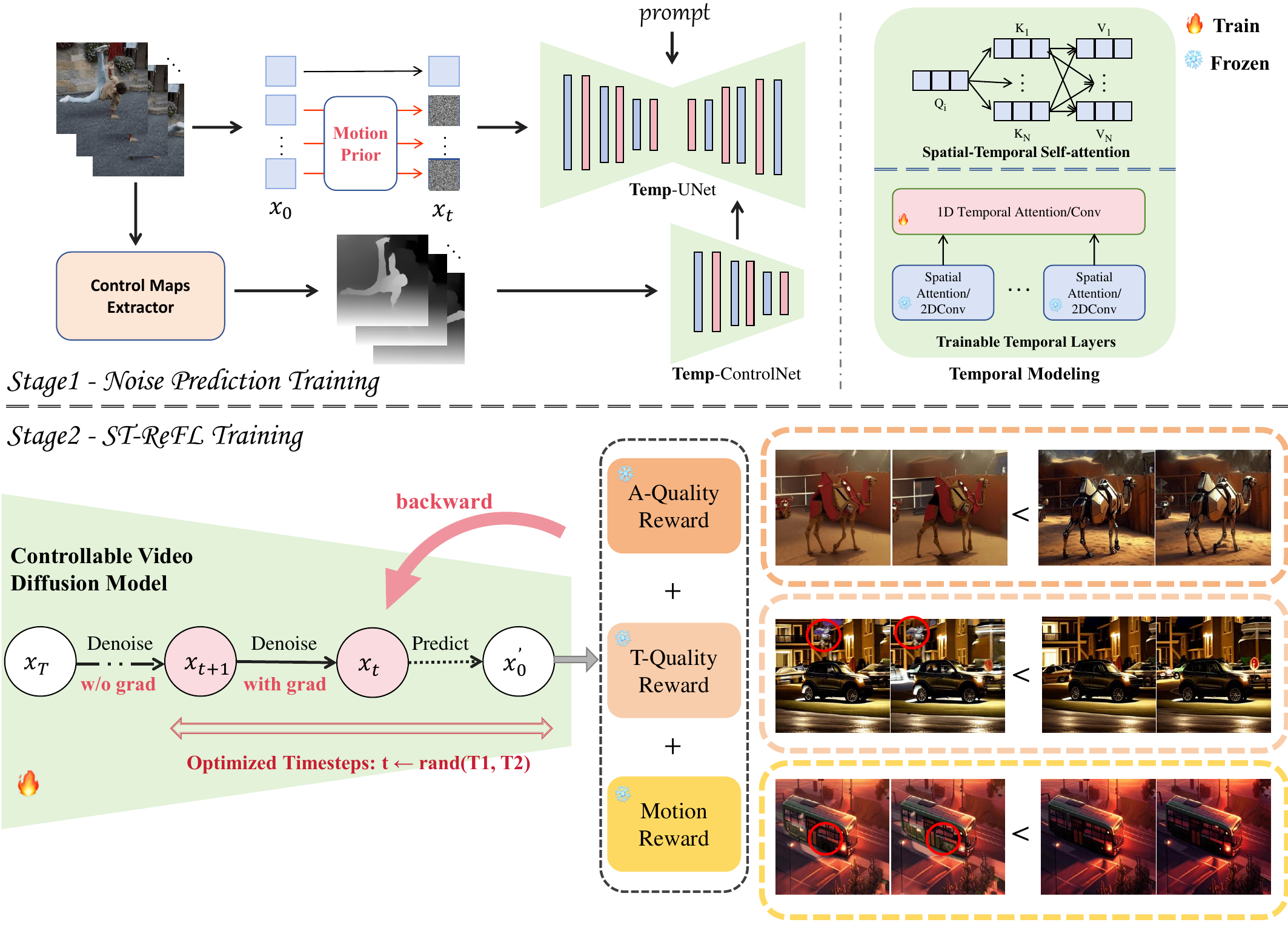}

   \caption{\textbf{Illustration of the Control-A-Video pipeline.} \textbf{Architecture:} Spatial-temporal self-attention and trainable temporal layers are applied to the UNet and ControlNet.(Temp denotes Temporal). \textbf{(1) Noise Prediction Training:} We add motion-aware noise to video latents (except the first frame) and train the model to predict the subsequent noise conditioned on the first frame, control maps, and text prompt. \textbf{(2)ST-ReFL Training:} A reward feedback loop optimizes the video diffusion model using (Aesthetic and Technical) Quality and Motion Rewards. It involves sampling $x_T$, denoising to $x_{t+1}$ without gradients, where $t \in [T_1,T_2] $, then denoising $x_{t+1}$ to $x_t$ with gradients to predict $x_0'$, which is decoded and scored by the reward models. The optimization aims to enhance aesthetic and technical quality while ensuring temporal consistency.}
   
   \label{fig:pipeline}
   \vspace{-0.2cm}
\end{figure*}
\vspace{-0.3cm}
\section{Method}

In this section, we will first introduce the controllable video diffusion model and its training and inference process with the first-frame condition scheme. Next, we introduce the motion-adaptive noise prior that help to improve the frame consistency. At the last section, we illustrate the reward feedback learning for video diffusion model that further enhances the capability of our model. 

\subsection{Video Diffusion Model with Control}
\label{architecture}
\textbf{Model Architecture:} Our model is built upon image diffusion model LDMs \cite{Ho_Jain_Abbeel_2020} and ControlNet \cite{zhang2023adding}, where both extra trainable motion layers are applied to UNet and ControlNet for temporal modeling. Similar to other work\cite{singer2022make, esser2023structure}, we add an additional 1D temporal layer to each 2-dimensional (2D) layer, including convolution and attention layers. To further enhance frame modeling, we employ a spatial-temporal self-attention mechanism, where spatial and temporal relationships are jointly model to capture dependencies across frames. 
As shown in Figure \ref{fig:pipeline}, each frame's are processed through either a 2D convolution layer or a spatial attention layer. Subsequently, these frame-level features are collectively passed to a trainable 1D convolutional layer or temporal attention for frame modeling. Moreover, to enable fine-grained modeling, we adapt the spatial self-attention mechanism by incorporating spatial-temporal self-attention across frames, which can be formulated as:

\begin{equation}\label{eq:SelfAttn}
SelfAttn(Q, K, V ) = Softmax(\frac{QK^T}{\sqrt{d}})V ,
\end{equation}
\begin{equation}\label{eq:KQV}
Q = W^Q \bar{v}_i, K = W^K [\bar{v}_0,...,\bar{v}_{N-1}], V = W^V [\bar{v}_0,...,\bar{v}_{N-1}]
\end{equation}
where $\bar{v}_i$ denotes the token sequence of frame $i$, and $[\bar{v}_0,...,\bar{v}_{N-1}]$ denotes the concatenation of the $N$ frames.
As shown in Eq. \ref{eq:KQV}, we concatenate features $K,V$ of $N$ frames so that each position has a global perception of all video frames and tends to generate more consistent results.





\label{training}
\textbf{Training with content prior:} To inherit the knowledge from image domain, we propose to introduce first frame as content prior to help generate more generalized videos. Specifically, as shown in the upper part of Figure \ref{fig:pipeline}, we add noise to each frames except the first one, and the model learns to denoise the noise with the content prior. This approach reduces the need for the model to memorize video content in the training set and instead focuses on learning to reconstruct motion, which makes it possible to achieve better results with fewer training resources. We thus train the model with the formulated loss function: 
\begin{equation}
min_\theta || \epsilon - \epsilon_\theta(x_t, t, c_p, c_f, \mathcal{E}(v^1)) ||_2^2
\end{equation}
, where $\epsilon$ is the ground truth noise and the diffusion model $\theta$ predict the noise with input $x_t$ at time step $t$ based on conditions prompt $c_p$, control maps $c_f$ and the first frame $v^1$. This simple yet effective strategy not only allows the model learns to effectively utilize the motion information from the control maps and follow the content from the first frame, but also able to auto-regressively generate longer videos. 

\textbf{T2I-I2V inference:} We generate the initial frame, $v^1$, by providing the model with gaussian noise of a single frame $x^1$ along with conditioning factors including a text prompt $c_p$ and a first frame control map $c_f^1$:
\begin{equation}
v^1 = ControlT2I(x^1 , c_p, c_f^1).
\end{equation}
Then we generate the following frames from gaussian noise $x$, where the first frame latent $x^1$ is $\mathcal{E}(v^1)$ from T2I model:
\begin{equation}
v = ControlT2V(x , c_p, c_f, \mathcal{E}(v^1)).
\end{equation}
With our proposed method of first-frame conditioning, our model is capable of generating video sequences with greater diversity than what is present in the training data. Additionally, our model has a distinct advantage in creating longer videos by utilizing previously generated frames as the initial frame in the subsequent iteration. This allows us to use an auto-regressive approach to produce longer videos.

\textbf{Classifier-Free Guidance}:
Based on typical Classifer-free guidance \cite{Ho_Salimans}, we incorporate a sampling strategy in \cite{esser2023structure} that treats noise prediction of video generated frame-by-frame as a negative representation needed to be avoided. Consequently, the final prediction of noise is calculated as:
\begin{equation}
\begin{aligned}
\hat\epsilon_\theta(x_t, t, c_p, c_f) = &\epsilon_{\theta I}(x_t, t, \emptyset, c_f)+\omega_v(\epsilon_\theta(x_t, t, \emptyset, c_f) - \epsilon_{\theta I}(x_t, t, \emptyset, c_f)) \\
&+\omega_t(\epsilon_\theta(x_t, t, c_p, c_f) - \epsilon_\theta(x_t, t, \emptyset, c_f)).
\end{aligned}
\label{eqn:equation}
\end{equation}
Here, $\omega_v$ and $\omega_t$ denote the scales of video guidance, $\emptyset$ denotes a null-text prompt, and $\epsilon_{\theta I}(x_t, t, \emptyset, c_f)$ denotes the prediction that each video frame is independently predicted. Just as a larger $w_t$ can enhance text guidance, a larger $w_v$ will result in a smoother overall effect.


\subsection{Motion-adaptive Noise Prior}
\label{noise_init}
\begin{wrapfigure}{r}{0.49\textwidth}
\vspace{-23pt}
\setlength{\belowcaptionskip}{-19pt}
    \centering
    \includegraphics[width=0.48\textwidth]{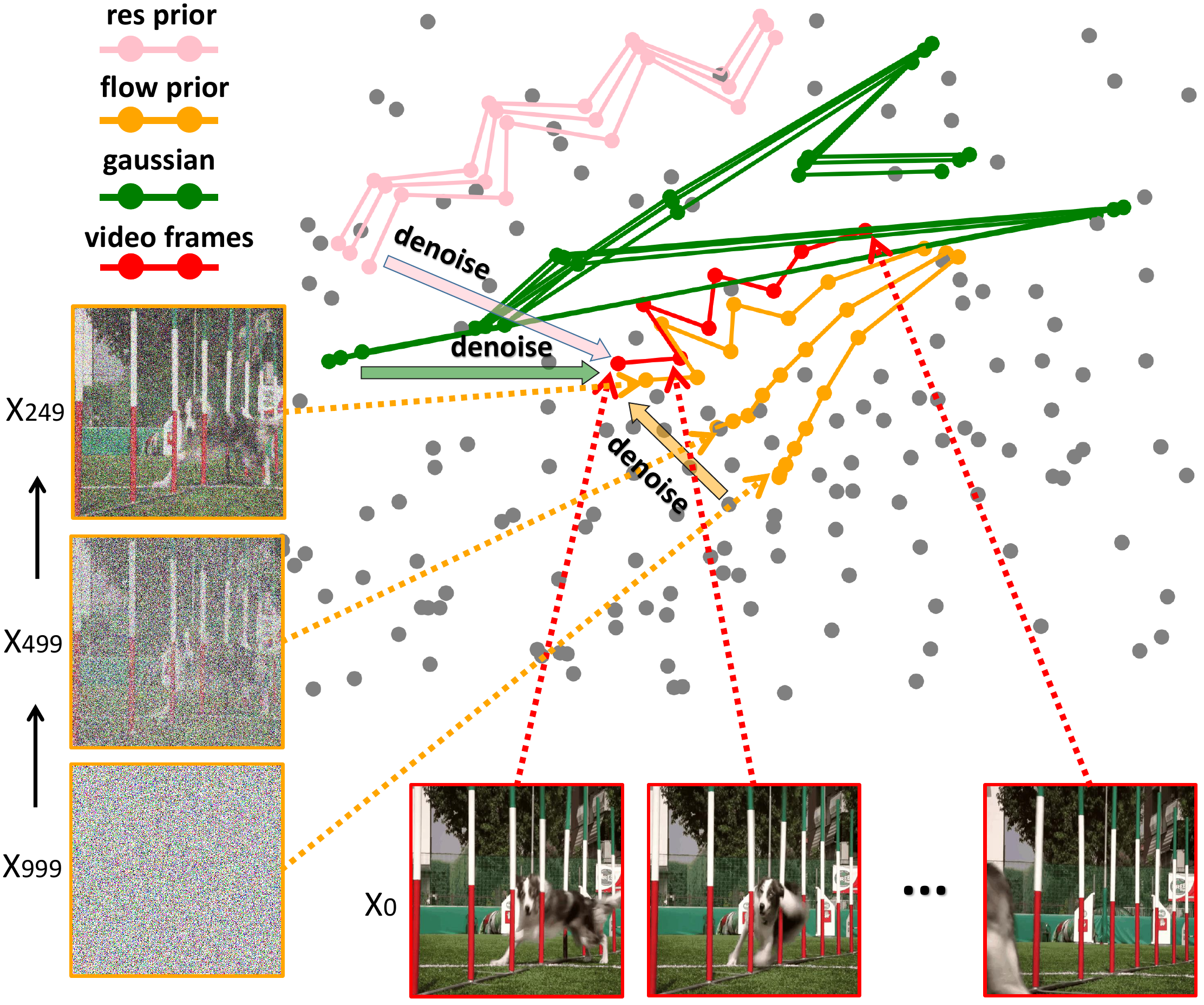}
    \caption{\textbf{Motion-adaptive Noise Prior}: t-SNE plot of noisy latents for video frames. Red: original video $X_0$. Noise is gradually added from timesteps 0 to 999, with pink: residual noise, orange: flow noise, green: Gaussian noise. Adjacent frames are linked, showing (1) frame similarity and (2) flow/residual noise preserves the line structure better than Gaussian noise, indicating improved consistency.}
    \label{fig:vis_noise}
\end{wrapfigure}

The diffusion model aims to denoise a signal by learning from Gaussian noise, where different initial noise samples can yield different results. In our study, we leverage the components of T2I models for video generation and observe that the latent spaces of consecutive video frames exhibit high correlation.
In Figure \ref{fig:vis_noise}, we can see that consecutive frames (represented by red points) are close to each other. However, when we add per-frame typical Gaussian noise (represented by green points) to each frame, we observe that the distribution between frames is disrupted. Not only does the distance between frames increase, but the overall distribution across frames also becomes distorted. Motivated by these findings, we propose a strategy to incorporate motion into the noise initialization process, aiming to align the noise closer to the video frames. To achieve this, we introduce two types of motion prior for video generation: flow-based and residual-based. By integrating motion-based noise into the latent space of the video, we empirically observe that the distribution of each frame maintains its similarity and coherence. As depicted in Figure \ref{fig:vis_noise}, the distribution of the flow-based noise (represented by orange points) is most similar to the original video, while the residual-based noise (represented by pink points) also exhibits high correlation. Intuitively, learning to reconstruct the video from motion-based noise is more feasible compared to using Gaussian noise since the latent similarity is preserved.
Specifically, the proposed algorithm is outlined in Algorithm \ref{algorithm}.

\textbf{Residual-based Noise Prior:} To maintain consistent noise in static regions and introduce varying noise in moving regions, we employ a residual-based noise prior. By computing the pixel residual between consecutive frames, we initialize the noise distribution accordingly after downsampling. This approach ensures that unchanged areas exhibit the same noise, while changing areas possess distinct noise patterns. Additionally, a threshold is utilized to differentiate static and dynamic regions, providing control over the smoothness of the generated videos.

\textbf{Flow-based Noise Prior:} In order to align the generated video's flow with the motion depicted in the frames, we introduce a flow-based noise prior. This involves computing the optical flow between consecutive frames in pixel space, followed by downsampling the flow information to the latent space. By propagating the flow through subsequent noise latents, we align the noise patterns with the expected motion flow, resulting in visually coherent and realistic videos.

\begin{figure}[t]
  \centering
  \scriptsize
  \begin{minipage}[t]{0.48\textwidth}
    \begin{algorithm}[H]
      \caption{\small Motion-adaptive Noise Propagation}
\label{algorithm}
\textbf{Notations:} Initialize noise $x$ with $N$ frames, $x^n \leftarrow \mathcal{N}(0, I) , n=0,...,(N-1) $ 
Input Video $: v = [ v^n, n=0,...,(N-1) ] $ 
$R_{thres}$ : Threshold for residual change 


\textbf{residual-based}

\For{ i from 1 to N}{
$ res = norm( v^i - v^{(i-1)} ) $ \\
$ res_{mask} = res > R_{thres} $ \\    
$res_{mask}$ = $DownSample(res_{mask})$ \\
$x^i = [x^i - x^{(i-1)} ]* res_{mask} +  x^{(i-1)}$ \\
}

\textbf{flow-based}

\For{ i from 1 to N}{
$ flow^i_{pixel} = flow( v^i , v^{(i-1)} ) $ \\
$ flow^i_{latent} $ = $DownSample(flow_{pixel})$ \\
$ x^i = GridSample(x^{(i-1)}, flow^i_{latent})$ \\ 
}
    \end{algorithm}
  \end{minipage}
  \hfill
  \begin{minipage}[t]{0.49\textwidth}
    \begin{algorithm}[H]
      \caption{\small Spatial-Temporal Reward Feedback Learning for Video Diffusion Models}
\label{algorithm_rlhf}
\textbf{Notations:} Video diffusion model $w$, quality reward model $R_q$, motion reward $R_m$. Total scheduler time steps $T$, randomly sampled timestep $t \in [T1, T2] $. Input latent $x$, output latent $x'$. $\alpha$ is the learning rate. \\
t $\gets$ rand(T1,T2) \\ 
$x_T$ $\sim$ $\mathcal{N}(0, I)$ \\
with no grad:  \\
\For{ i from T to t+2}{ 
$x_{i-1}$ $\gets$ $w(x_i)$ \\
}
with grad: $x_{t}$ $\gets$ $w(x_{t+1})$ \\
$x_0'$ $\gets$ $x_t$  \quad  // predict \\
$v_0'$ $\gets$ $x_0'$, $v_0$ $\gets$ $x_0$   \quad // decode   \\
$L_{motion}$ $\gets$ $R_{m}(v_0, v_0')$, $L_{quality} \gets R_{q}(v_0')$ \\
$\mathcal{L}_{ST} \gets L_{motion} + L_{quality}$ \\
$\Delta \gets \frac{\partial \mathcal{L}_{ST}}{\partial w}$ \\
$w'$ $\gets$ w - $\alpha$ $\Delta$  \quad // update 
    \end{algorithm}
  \end{minipage}
  \vspace{-0.5cm}
\end{figure}








\subsection{Reward Feedback Learning for Controllable Video Generation}


To enhance both the quality and motion consistency of the generated videos, we seamlessly integrate Reward Feedback Learning\cite{imageward} into our video diffusion model. As illustrated in the lower part of Figure \ref{fig:pipeline}, our training procedure entails the sampling of $x_T$, followed by denoising it to timestep $(t+1)$ without gradients. Subsequently, we denoise from $x_{t+1}$ to $x_t$ with gradient and predict $x_0'$. It is essential to highlight that selecting a small value for $t$ yields a more precise approximation of $x_0'$ and therefore leads to superior performance. Upon obtaining the predicted video, we assess its quality through a dual-pronged approach, considering both motion consistency and overall video quality.

To quantify the motion consistency of generated videos, we introduce the use of residual and optical flow~\cite{raft_flow} among frames as a motion field. The motion reward is then computed as the negative difference between this motion field and the motion field of the input video. The loss function for motion consistency is formulated as:
\begin{equation}
L_{motion} = - \lambda_{mr} \cdot R_{mr}(v, v') - \lambda_{mf} \cdot R_{mf}(v, v'),
\end{equation} 
where the ground-truth video is denoted as $v$, and $v'=Decoder(x_0')$ representing the predicted video. The residual-based motion reward $R_{mr}$ and flow-based motion reward $R_{mf}$ are calculated and subsequently combined with a weighted sum, contributing to the overall loss.

Another pivotal metric for video generation involves enhancing both the technical and aesthetic qualities in the output. Technical quality encompasses distortions, such as over-exposure, noise, and blur, present in the generated frames. We evaluate technical quality using the MUSIQ metric\cite{musiq} as a measure of technical quality reward.
Additionally,  to enhance aesthetic quality, we employ a reward model\cite{imageward} trained on a human-rated dataset, thereby enriching the overall aesthetic quality of the generated video. Consequently, the loss for quality is expressed as weighted sum of two quality reward:
\begin{equation}
L_{quality} = \lambda_{qt} \cdot ReLU( b_{qt} - R_{qt}( v' ) ) + \lambda_{qa} \cdot ReLU( b_{qa} - R_{qa}( v' ) ) .
\end{equation} 
Here, we compute frame-wise technical reward $R_{qt}$ and aesthetic reward $R_{qa}$, where $b_{qt}$ and $b_{qa}$ represent the upper boundaries of the reward models. $\lambda_{qt}$ and $\lambda_{qa}$ serve as weights.

We thus drive a spatio-temporal reward loss in sum of the above two losses as:
\begin{equation}
\mathcal{L}_{ST} = L_{motion} +  L_{quality}.
\end{equation} 
Then we update the model via feedback learning, as illustrated in Figure \ref{fig:pipeline}. For a more detailed description, the ST-ReFL algorithm is presented in Algorithm \ref{algorithm_rlhf}.

\section{Experiments}
\label{exp}

\subsection{Implementation Details}
\textbf{Dataset Settings:}~ We collect 0.1M video clips from films\protect\footnotemark[1] and caption them with BLIP\cite{BLIP} and 0.1M image-text pairs randomly sampled from LAION \cite{schuhmann2022laion5b}. 

\footnotetext[1]{Note that our model can be trained on Webvid\cite{webvid} as well, but it contains watermarks.}

\noindent\textbf{Training Settings:}~ Our training regimen commences with the fine-tuning of Stable Diffusion v1.5\cite{Ramesh_Dhariwal_Nichol_Chu_Chen}, incorporating DiffusionDB\cite{diffusiondb} and a curated dataset. This refinement aims to improve the Text-to-Image (T2I) model, which is used for initialization for Temp-UNet. The initialization of Temp-ControlNet stems from ControlNet\cite{zhang2023adding}. We train the temporal layers with mixture of video and image data. The configuration includes a resolution set at $512\times 512$, a batch size of $16$, a learning rate of $10^{-5}$, and a total of $20k$ training steps. We demonstrate the architecture with three control types: Canny edge maps\cite{Canny_2009}, HED edge maps\cite{Xie_Tu_2016}, and depth maps\cite{Ranftl_Lasinger_Hafner_Schindler_Koltun_2020}. After the first stage training, we further finetune the model with reward feedback learning, where the hyper-parameters are set as $\lambda_{mr}=1.0, \lambda_{mf}=0.02, \lambda_{qt}=0.001, b_{qt}=100, \lambda_{qt}=0.005, b_{qa}=2,$. We use DDIM to sample 20 steps and set $T_1=0, T_2=5$ during ST-ReFL training.

\noindent\textbf{Inference Settings:}~ Text guidance is scaled at 7.0 and video guidance adopts a scale of 1.5. We leverage 20 sampling steps with DDIM to speedup inference. The initialization threshold is set at 0.1 for residual-based noise prior.

\noindent\textbf{Evaluation Settings:}~ We conduct comprehensive experiments to demonstrate the effectiveness of our proposed strategies qualitatively and quantitatively. 
In the quantitative comparison, we adopt depth maps from 30 video clips from Davis \cite{davis_2017}, which are used to generate videos based on a given text prompt.  
Following previous works~\cite{videocomposer}, we calculate the cosine similarity between video embeddings of the output videos and text embeddings of the given prompts with CLIP \cite{clip2021}\protect\footnotemark[2] to evaluate text alignment. 
To assess semantic consistency between different frames, we measure the similarities for frame CLIP embeddings of output videos. 
Aiming at reflecting consistency with control signals, we compute the depth map errors between input and output videos.  
Inspired by VBench, we adopt MUSIQ to evaluate frame quality objectively, targeting the distortion (e.g., over-exposure, noise, blur)presented in the generated frames. 
Furthermore, we test our model on two subjective aesthetic metrics, ImageReward~\cite{imagereward} and aesthetic-predictor\protect\footnotemark[3].
\footnotetext[2]{https://huggingface.co/openai/clip-vit-large-patch14}
\footnotetext[3]{https://github.com/LAION-AI/aesthetic-predictor}
\vspace{-0.3cm}
\subsection{Main Results}


\subsubsection{Controllable Video Generation.}
We showcase three types of controls extracted from video to demonstrate our system's capacity to generate videos conditioned on various control types. 
As shown in Figure \ref{fig:head}, we found that depth maps provide less structural information than edge maps, resulting in more diverse video outputs. 
Edge maps, on the other hand, produce videos with enhanced details. 
For instance, in the first row of Figure \ref{fig:head}, we transform a human into a cartoon robot using depth map control. 
In contrast, using edge maps in the third row still results in a human-robot transformation, but with more intricate details retained, such as the clothing pattern.

\begin{figure}[!t]
   \centering
   \includegraphics[width=0.99\linewidth]{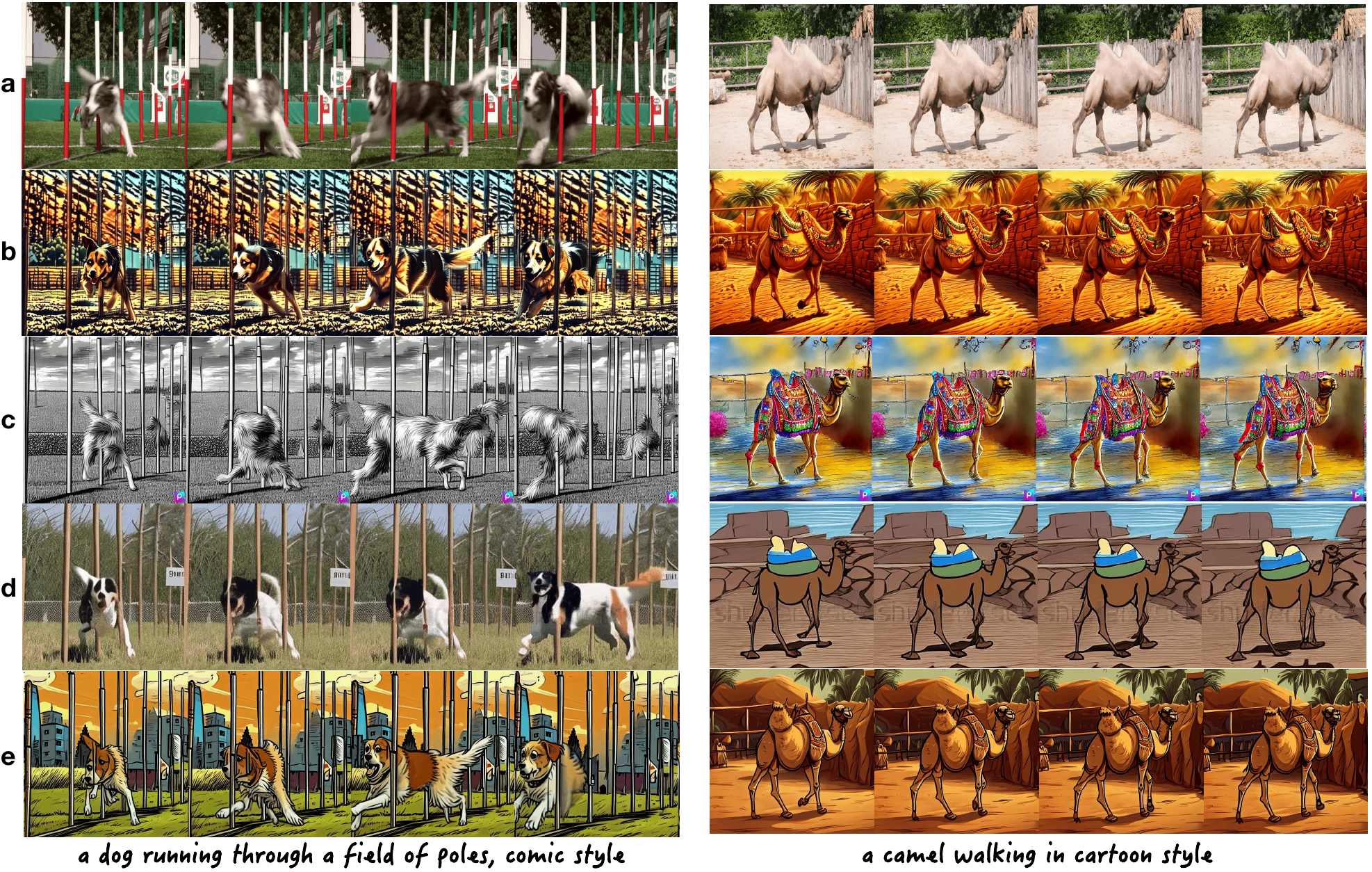}
       \caption{\textbf{Qualitative Comparison:} (a) Input video, (b) Animatediff, (c) Text2Video-Zero, (d) Videocomposer, (e) Ours. We showcase a challenging scenario of a fast-moving dog(left) and slow-moving(right) camel. Compared to other models, our method demonstrates superior performance in generating high-quality, temporally consistent results that accurately align with the given text prompt. }\label{fig:comp4}
\vspace{-0.3cm}
\end{figure}

\begin{table}[t]
\centering
\caption{Comparison with other models. ``Img-Re'' is short for ImageReward and ``Aes-pred'' refers to Aesthetic-predictor.}
\small
\begin{adjustbox}{max width=\textwidth}
\begin{tabular}{l|cc|c|ccc|ccc}
\toprule
\multirow{2}{*}{\textbf{Model}} & \multicolumn{2}{c|}{\textbf{CLIP Metrics} $\uparrow$} & \textbf{Map Errors} $\downarrow$ & \multicolumn{3}{c|}{\textbf{Frame Quality} $\uparrow$} & \multicolumn{3}{c}{\textbf{User Study} $\uparrow$} \\
                       & prompt      & frames      & depth        & MUSIQ & Img-Re & Aes-pred & Text-Align & Consistency & Quality \\ \midrule
\textbf{Text2video-Zero~\cite{text2video-zero1}}        &       0.281        &       \textbf{0.964}        &    0.110        &   71.0    &       0.531      &        5.38          &     3.4       &     3.0        &     2.8    \\
\textbf{VideoComposer~\cite{videocomposer1}}           &       0.272           &       0.937       &      0.149           &    52.8   &       -0.002      &           4.60          &      3.1      &       3.6      &    2.9     \\
\textbf{AnimateDiff~\cite{animatediff1}}            &          0.274         &        0.958            &          0.097        &     59.6   &      0.67       &          \textbf{5.79}           &      3.8      &      4.0       &  4.0        \\
\textbf{Control-A-Video}      &  \textbf{0.290}          & 0.959      & \textbf{0.089}              & \textbf{73.3}          & \textbf{1.03} &  5.73                   &     \textbf{4.4}       &       \textbf{4.5}      &       \textbf{4.2}  \\ \bottomrule
\end{tabular}
\end{adjustbox}
\label{tab:main_result}
\vspace{-0.4cm}
\end{table}


\subsubsection{Qualitative Analysis.}
\footnotetext[1]{We use our fine-tuned T2I model as the initialization of AnimateDiff.}
Control-A-Video can incorporate various control signals, such as depth maps, enabling it to naturally generalize to video editing and video style transfer. 
We conduct qualitative comparison on video editing with other three typical models, including Text2video-Zero~\cite{text2video-zero1} performing zero-shot video editing, VideoComposer~\cite{videocomposer1} which is a controllable video generation method, and AnimateDiff \cite{animatediff1}\protect\footnotemark[1] for animating personalized T2I models. 
As shown in Figure \ref{fig:comp4}, the first row depicts the original video, from which we extract depth maps and use new prompts to create a style transfer video. 
In the second row, although the results of AnimateDiff have consistent subjects, the background in the first case has too much noise, which affects the visual quality. 
The third row exhibits the results of Text2Video-Zero, which fails to create a correct style and the dog in the first case is blurry. 
As for Videocomposer, it can't maintain consistency of the dog in the first case and its style is unrelated. 
In the last row, we present the videos generated by our model, which is most clear, consistent and text-aligned. 
More Comparison will be shown in supplementary material.

\begin{figure}[!t]
   \centering
   \includegraphics[width=\linewidth]{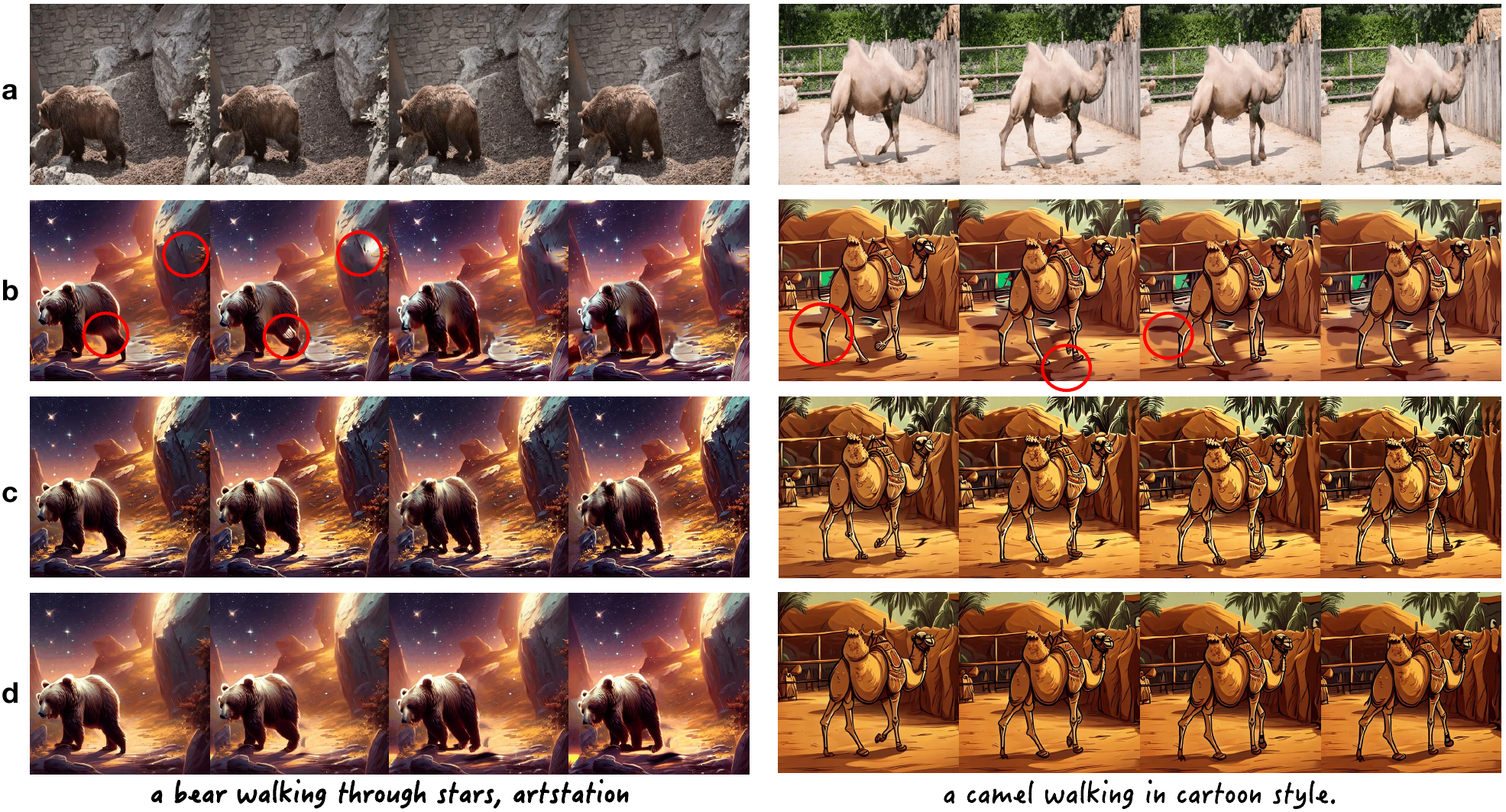}
       \caption{ \textbf{ Ablation study of motion-adaptive noise prior.} (a) Input video. (b) Without motion prior, exhibiting flickering and artifacts. (c) With Optical Flow-based prior, maintaining consistency. (d) With Residual-based prior, stable background and coherent subject motion.}
       \label{fig:noise}
       \vspace{-0.3cm}
\end{figure}

\begin{table}[!t]
\centering
\caption{Quantitative ablation comparison for our noise initialization strategies in the first training stage. The baseline indicates the model trained with depth maps as control. RNI and FNI correspond to pixel residual-based and optical flow-based noise initialization.}
\begin{tabular}{l|cc|c|ccc}
\toprule
\multirow{2}{*}{\textbf{Model}} & \multicolumn{2}{c|}{\textbf{CLIP Metrics} $\uparrow$} & \textbf{Map Errors} $\downarrow$ & \multicolumn{3}{c}{\textbf{Frame Quality} $\uparrow$}           \\
                                & prompt               & frames              & depth               & MUSIQ         & Img-Re    & Aes-pred \\ \midrule
\textbf{Baseline}               & 0.278                & 0.950               & 0.090               & \textbf{70.2} & 0.593          & 5.41                \\
\textbf{+RNI}                   & 0.284                & 0.952               & 0.089               & 69.3          & 0.737          & 5.49                \\
\textbf{+FNI}                   & \textbf{0.288}       & \textbf{0.953}      & \textbf{0.086}      & 67.1          & \textbf{0.787} & \textbf{5.52}       \\ \bottomrule
\end{tabular}
\label{tab:noise_ablation}
\vspace{-0.3cm}
\end{table}

\vspace{-0.3cm}
\subsubsection{Quantitative Comparison.} 
We conduct comparison with other end-to-end models on the 30 video clips from Davis. 
As shown in Table \ref{tab:main_result}, our model outperforms previous models in four out of six metrics. Control-A-Video achieves the highest prompt clip score and exhibits the lowest depth map errors, demonstrating superior text alignment and video consistency. Especially in terms of frame quality, our model realizes an improvement of +2.3 and +0.36 points over the previous best models on the MUSIQ and ImageReward scores, respectively. The aesthetic predictor metric of our model is comparable to that of AnimateDiff. Note that Text2video-Zero secures a higher frames clip score than ours, likely attributable to its videos having smaller ranges of movement.

As for user study, 20 participants were surveyed to evaluate the textual alignment, consistency and visual quality of the generated videos by utilizing a rating scale ranging from 1 to 5. 
Our model achieves the best performance in human evaluation.




\vspace{-0.3cm}
\subsection{Ablation Studies}
\label{sec:ablation}

\subsubsection{Motion-adaptive Noise Prior.}
\label{ablation noise}
To demonstrate that pixel residual-based and optical flow-based noise initialization (simplify as RNI and FNI) strategies boost video consistency, we perform visualization case study and quantitative comparison in the first training stage. 
Figure~\ref{fig:noise} illustrates the effect of different noise initialization strategies. 
In the second row, the bear in the first case has artifacts and the backgrounds in both cases have many flickers. 
In contrast, due to the presence of motion-adaptive noise prior, the video in last two rows are more consistent, which exhibits superior visual performance. 

As summarized in Table~\ref{tab:noise_ablation}, both noise initialization methods, RNI and FNI, improve all metrics except MUSIQ over the baseline. 
The results demonstrate that videos initialized with RNI and FNI exhibit higher prompt consistency and frame-level consistency. 
Additionally, errors in depth maps are reduced, indicating that the motion prior successfully contribute to maintaining the structure of the source video. 
Notably, in terms of aesthetic metrics, RNI and FNI enhance the ImageReward score by 0.144 and 0.194, respectively. 
The ablation study for the threshold of residual-based noise initialization is detailed in the Appendix.





\begin{figure}[!t]
   \centering
   \includegraphics[width=\linewidth]{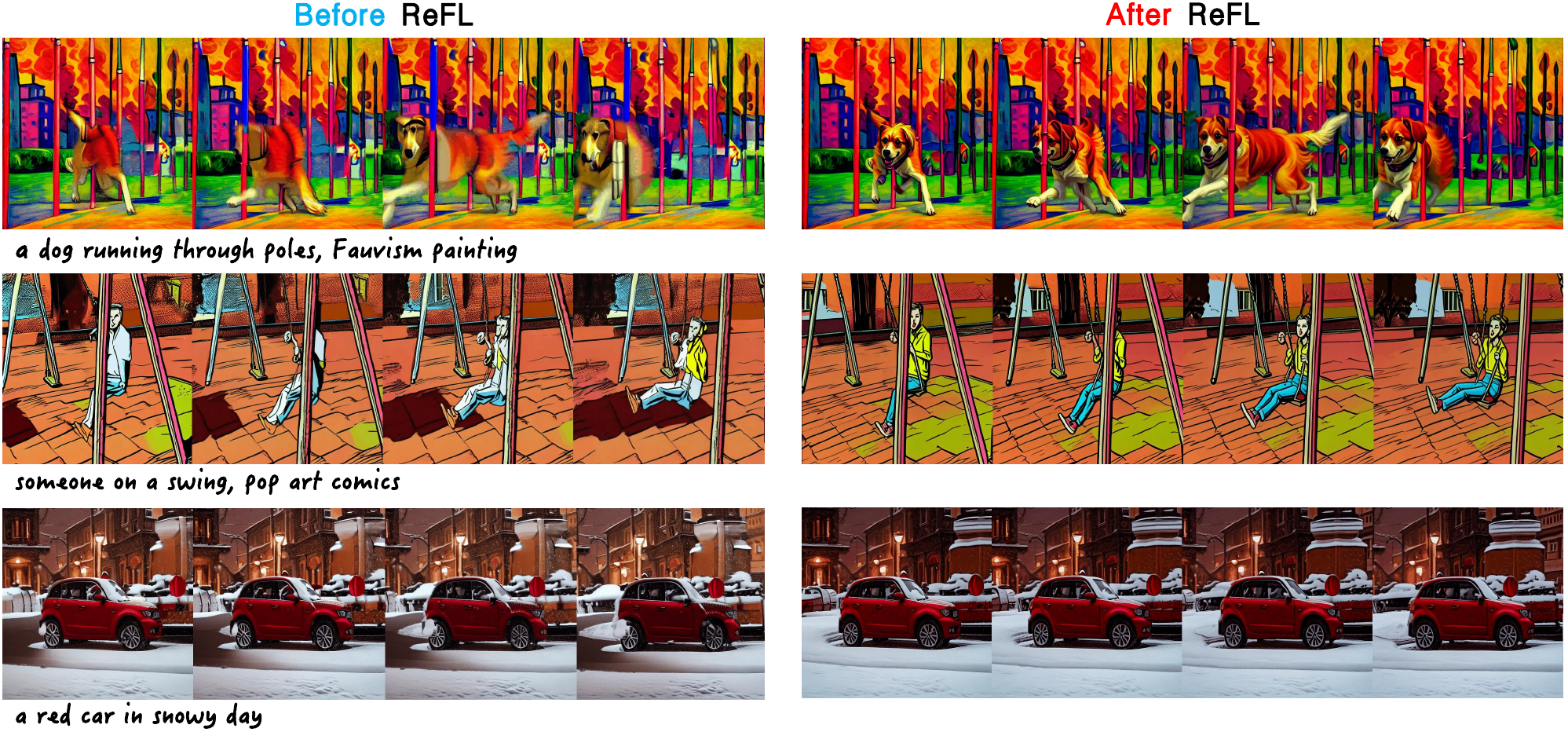}
        \caption{\textbf{Qualitative ablation study of ST-ReFL optimization}. Three examples before and after ST-ReFL show improved consistency and quality through feedback tuning from motion and quality reward models.}
       \label{fig:rlhf_compare}
       \vspace{-0.3cm}
\end{figure}

\subsubsection{Spatial-Temporal Reward Feedback Learning.}
We integrate four reward feedback signals into controllable video generation, two of which are designed to improve video quality and the other two for enhancing consistency. 
As depicted in Figure \ref{fig:rlhf_compare}, we provide a qualitative comparison for the model with and without ST-ReFL training. 
Take the first row as an example, the dog has clear texture and its facial features appear lifelike after ST-ReFL training, indicating ST-ReFL improves frame quality and aesthetic effect. 
In the second row, before ST-ReFL training, there are some color changes on the floor and the clothes of the person. 
In the third rows, there are some artifacts on the car wheel. 
Nonetheless, our final model delivers consistent videos of high quality, as evidenced by the results.

To investigate the individual impact of each reward signal, we conduct quantitative ablation experiments during the second stage training, systematically omitting one signal at a time. 
The results are shown in Table \ref{tab:ablation_reward}. 
Compared to the results of the first four lines, our complete model achieves a balance across all metrics and gains the best overall performance. 
Specifically, as observed in the first row of the table, the absence of the aesthetic reward compromises the model’s performance on video clip metrics and frame-level aesthetic evaluation. 
The second line highlights the results without technical reward are the worst on MUSIQ, which indicates technical reward plays an important role in frame quality. 
In the third line, when removing residual reward, the depth error and frame quality get worse. 
Comparing results of the last two lines, optical flow reward makes a significant improvement on motion consistency and ImagReward score.


\begin{table}[t]
\centering
\caption{Ablation studies of different reward signals. The best scores are in \textbf{bold} and the second best scores are \ul{underlined}.}
\label{tab:ablation_reward}
\begin{adjustbox}{max width=\textwidth}
\small
\begin{tabular}{l|cc|c|ccc}
\toprule
\multirow{2}{*}{\textbf{Model}}  & \multicolumn{2}{c|}{\textbf{CLIP Metrics} $\uparrow$} & \textbf{Map Errors} $\downarrow$ & \multicolumn{3}{c}{\textbf{Frame Quality} $\uparrow$}          \\
                                 & prompt               & frames              & depth                & MUSIQ         & Img-Re   & Aes-pred \\ \midrule
\textbf{w/o Aesthetic reward}    & 0.288                & 0.955               & {\ul 0.089}           & {\ul 74.4}    & 0.897         & 5.70                \\
\textbf{w/o Technical reward}    & 0.289                & {\ul 0.958}         & \textbf{0.086}          & 70.7          & 0.983         & 5.62                \\
\textbf{w/o Residual reward}     & \textbf{0.292}       & {\ul 0.958}         & 0.093                  & 72.4          & {\ul 0.996}   & 5.69                \\
\textbf{w/o Optical flow reward} & 0.286                & 0.957               & 0.094                  & \textbf{75.6} & 0.866         & \textbf{5.75}       \\
\textbf{Control-A-Video (Ours)}  & {\ul 0.290}          & \textbf{0.959}      & {\ul 0.089}              & 73.3          & \textbf{1.03} & {\ul 5.73}          \\ \bottomrule
\end{tabular}
\end{adjustbox}
\vspace{-0.4cm}
\end{table}

\vspace{-0.3cm}
\section{Conclusion}

In this paper, we introduce a controllable Text-to-Video (T2V) framework for generating videos based on text prompts and control maps. Our framework leverages content prior extracted from an initial frame, facilitating the transition from the image to video domain. Additionally, we enhance temporal consistency by employing motion prior derived from residual/optical flow initialization. Moreover, our Spatio-Temporal Reward Feedback Learning algorithm optimizes the model with dedicated reward models, resulting in significant improvements in both video quality and motion coherence. Our combined strategies offer innovative avenues for enhancing the quality and controllability of diffusion-based video synthesis through prior initialization and reward feedback learning.


\clearpage
\setcounter{page}{1}

\appendix
\section*{\centering \LARGE Appendix} 

\section{Preliminary}

\textbf{Diffusion Models}: Given an input signal $x_0$, a diffusion forward process is defined as:
\begin{equation}
p_\theta(x_{t} | x_{t-1}) = \mathcal{N}(x_t ; \sqrt{1-\beta_{t-1}}x_{t-1}, \beta_t I), ~~~t = 1, ..-., T
\end{equation}

where $T$ is the total timestep of the diffusion process. A noise depending on variance $\beta_t$ is gradually added to $x_{t-1}$ to obtain $x_t$ at the next timestep and finally reach $x_T \in \mathcal{N}(0, I)$. The goal of the diffusion model \cite{Ho_Jain_Abbeel_2020} is to learn to reverse the diffusion process (denoising). Given a random noise $x_t$, the model predicts the added noise at the next timestep $x_{t-1}$ until the origin signal $x_0$. 

\begin{equation}
p_\theta(x_{t-1} | x_t) = \mathcal{N}(x_{t-1}; \mu_\theta(x_t,t), \Sigma_\theta(x_t,t) ), ~~~t = T, ..., 1
\end{equation}

We fix the variance $\Sigma_\theta(x_t,t)$ and utilize the diffusion model with parameter $\theta$ to predict the mean of the inverse process $\mu_\theta(x_t,t)$. The model can be simplified as denoising models $\epsilon_\theta(x_t, t)$, which are trained to predict the noise of $x_t$ with a noise prediction loss:
\begin{equation}
min_\theta || \epsilon - \epsilon_\theta(x_t, t, c_p) ||_2^2
\end{equation}

where $\epsilon$ is the added noise to the input image $x_0$, the model learn to predict the noise of $x_t$ conditioned on text prompt $c_p$ at timestep $t$.

\noindent\textbf{Latent diffusion Models}: LDM propose to apply a compressed latent code $z$ rather than the image signal $x$ in diffusion process to speed up the denoising process. The image $x$ is encoded by an encoder $\mathcal{E}$ to obtain the latent code $z=\mathcal{E}(x)$, and the model learns to denoise in latent space. During inference, the reconstructed latent code $z_0$ can be reconstructed by a decoder $\mathcal{D}$, $x_0=\mathcal{D}(z_0)$ to obtain the generated image.

\noindent\textbf{ControlNet}: ControlNet is a neural network architecture that enhances pretrained image diffusion models with task-specific conditions by utilizing trainable layers copied from the original diffusion model. These layers are then fine-tuned based on specific control maps such as edge, depth, and segmentation inputs. The loss with additional control can be formulated as:
\begin{equation}
min_\theta || \epsilon - \epsilon_\theta(x_t, t, c_p, c_f) ||_2^2
\end{equation}

where the control map $c_f$ is an additional control. Our research draws inspiration from ControlNet and expands its application into video synthesis. In the case of a video, the input signal $x$ and control $c_f$ is extended to a sequence of N frames.

\section{More Experiment results}
\subsection{Cases show}
Please refer to the uploaded video in the attachment, which includes a majority of the videos referenced in this paper, along with additional cases. 
The video can also be downloaded from this \href{https://drive.google.com/file/d/1Bvvg-Nq67Oi4JavYiJOOcUtwbE5SD5g_/view?usp=sharing}{\textcolor{blue}{link}}. More results please refer to the supplementary materials, including the videos of ablation study and comparison, with the prompt as filename.

\textbf{(1) Ablation Study of Motion Prior and ST-ReFL}: We present the results of an ablation study evaluating our main contributions: motion prior and ST-ReFL. \textbf{Motion Prior:} As demonstrated, without the motion prior, the results tend to exhibit flickering issues. For instance, in the first case (the camel demo), the background becomes distorted without the motion prior, and the technical quality deteriorates.  \textbf{ST-ReFL:} We demonstrate that incorporating ST-ReFL significantly improves both the consistency and quality of the output videos. For example, in the second case (the dog demo), the clarity and visual appeal are noticeably enhanced compared to other methods, showcasing the effectiveness of our approach in generating high-quality and aesthetically pleasing video outputs.

\textbf{(2) Ablation Study of Threshold for Residual-based Motion Prior}: Once trained with residual-based noise, we can set different threshold to control the smoothness of noise prior. We showcase generated videos with different noise: identical noise (threshold=1.0), distinct noise (threshold=0.0), and motion-enhanced noise(threshold=0.1). In the video exhibited in the link, the threshold (0.1) we select effectively balances consistency and smoothness to produce satisfactory results.

\textbf{(3) Comparison with Other Methods}: We conduct a comprehensive comparison between our proposed method and other existing approaches in the field of video editing as Figure \ref{fig:comp4}. This analysis aims to highlight the unique advantages of our model and its potential for advancing the current state-of-the-art techniques.

\textbf{(4) Auto-regressive Generation}: We showcase the practical applications of our model in the domain of video editing. Leveraging our novel frame-conditioning strategy, these videos are generated using an auto-regressive approach, repeatedly extending the sequence two or three times. This auto-regressive capability emerges as an exciting byproduct of our innovative approach, unlocking new possibilities for extended video generation.

\textbf{(5) Support for Different Control Maps}: We provide more cases than Figure \ref{fig:head} to demonstrate the versatility of our model in supporting various control maps as conditions. Depth control offers enhanced flexibility, allowing for more creative freedom, while edge control ensures greater consistency in the edited videos.

\clearpage  

%
%
\bibliographystyle{splncs04}
\bibliography{main}
\end{document}